\pdfoutput=1

\documentclass[11pt]{article}

\usepackage{EMNLP2022}
\usepackage{times}
\usepackage{latexsym}
\usepackage{enumitem}

\usepackage[T1]{fontenc}

\usepackage[utf8]{inputenc}

\usepackage{microtype}

\usepackage{inconsolata}

\usepackage{graphicx}
\graphicspath{ {./images/} }

\title{DeepGen: Diverse Search Ad Generation and Real-Time Customization}

\author{Konstantin Golobokov$^{\diamondsuit}$, \hspace{0.25em} {\bf Junyi Chai$^{\clubsuit}$,} \hspace{0.25em} {\bf Victor Ye Dong$^{\clubsuit}$,} \\
{\bf Mandy Gu$^{\clubsuit}$,} \hspace{0.25em} {\bf Bingyu Chi$^{\clubsuit}$,} \hspace{0.25em} {\bf Jie Cao$^{\clubsuit}$,} \hspace{0.25em} {\bf Yulan Yan$^{\spadesuit}$,} \hspace{0.25em} {\bf Yi Liu$^{\clubsuit}$}\\
$^{\diamondsuit}$ Azure AI, $^{\clubsuit}$ Bing Ads, $^{\spadesuit}$ News \& Feeds \\ Microsoft, Redmond, USA \\ \texttt{\{FirstName.LastName\}@microsoft.com}}
\begin{document}
\maketitle
\begin{abstract}
We present DeepGen, a system deployed at web scale for automatically creating sponsored search advertisements (ads) for Bing Ads customers. We leverage state-of-the-art natural language generation (NLG) models to generate fluent ads from advertiser's web pages in an abstractive fashion and solve practical issues such as factuality and inference speed. In addition, our system creates a customized ad in real-time in response to the user's search query, therefore highlighting different aspects of the \textit{same} product based on what the user is looking for. To achieve this, our system generates a \textit{diverse} choice of smaller pieces of the ad ahead of time and, at query time, selects the most relevant ones to be stitched into a complete ad. We improve generation diversity by training a controllable NLG model to generate multiple ads for the same web page highlighting different selling points. Our system design further improves diversity horizontally by first running an ensemble of generation models trained with different objectives and then using a diversity sampling algorithm to pick a diverse subset of generation results for online selection. Experimental results show the effectiveness of our proposed system design. Our system is currently deployed in production, serving ${\sim}4\%$ of ads globally on Bing.
\end{abstract}

\section{Introduction}
\label{introduction}
Search advertising is the largest segment of digital advertising for its projected \$203B out of \$515B market share worldwide in 2022 \citep{statista}. Traditionally, advertisers manually create ads for their web pages to start an advertising campaign. There is a growing need to automate this process, either to lessen the burden for small and medium businesses, or to create millions of ads for large businesses that have lots of products. 

\begin{figure}[h]
\includegraphics[width=\columnwidth]{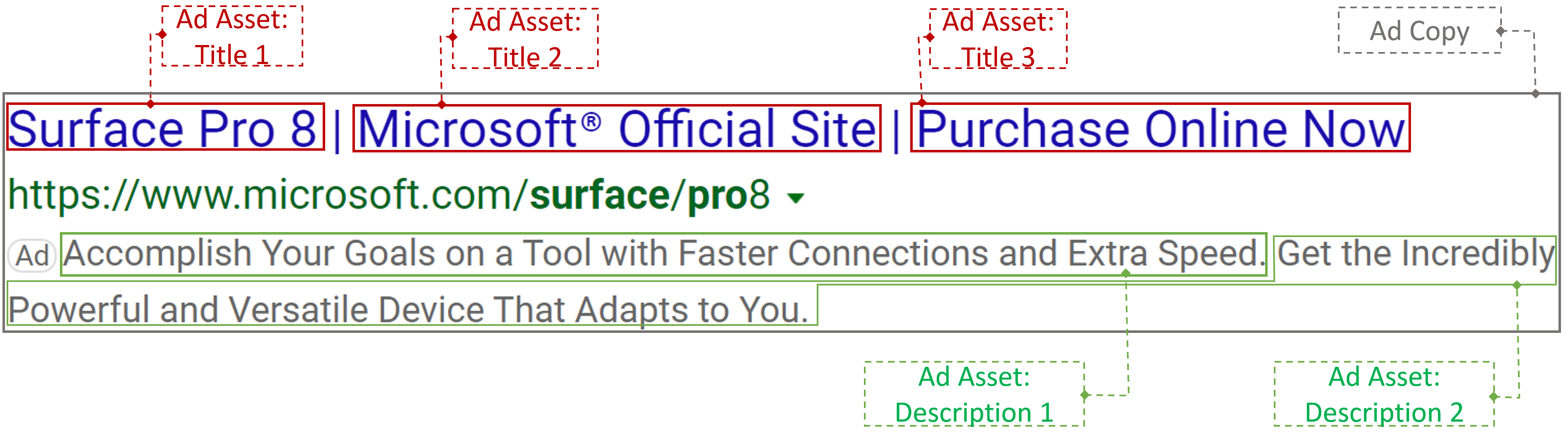}
\caption{An example of an ad copy (grey box) comprised of ad assets. Red box is used for ad title assets, and green box is used for ad description assets. This ad could be shown for search query ``Surface 8".}
\label{fig:AdComponentsXbox}
\end{figure}

A classical automated ad generation system relies on extraction rules as described in Section~\ref{sec:Models}, for example, extracting key phrases from advertiser's web pages as ad titles. However, per our experience, extraction-based methods are not very successful in generating the much longer ad description. Refer to Figure~\ref{fig:AdComponentsXbox} for the example ad title and description assets. Therefore, we aim to generate ads in an abstractive fashion.  In this work, we focus on improving ad performance from two aspects: factuality and customization.

\begin{figure*}[ht]
\begin{center}
\centerline{\includegraphics[width=\textwidth]{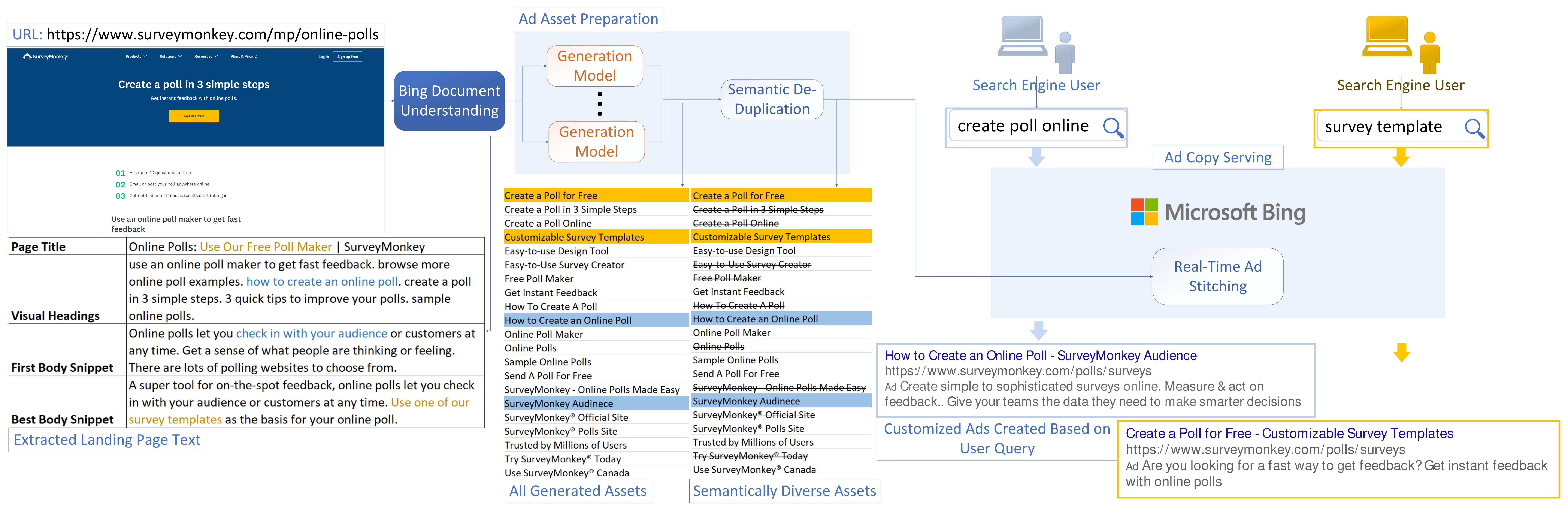}}
\caption{An illustration of the end-to-end DeepGen system. First, multiple ad assets are generated based on various parts of the advertiser's web page. Semantically diverse ad assets are then selected and prepared for serving. Finally, customized ads are created based on user queries. Transparent blocks are the NLP models, solid blocks are the surrounding infrastructure. Generative models are shown in orange, discussed in Section~\ref{sec:genMethodology}. The rest of the system is presented in Section~\ref{sec:SystemDesign}.}
\label{fig:DeepGenWorkflow}
\end{center}
\end{figure*}

To achieve the optimal ad performance, our current system creates a customized ad in real-time in response to a user's search query. As shown in Figure~\ref{fig:DeepGenWorkflow}, different ads are displayed for different queries, although they are advertising the same web page. We dynamically customize ad copies by stitching the generated ad assets together given the user's search context, approximating the ultimate goal of real-time customized generation. Our work makes the following contributions:
\begin{enumerate}[noitemsep]
    \item We demonstrate an NLG application that leverages cutting-edge models, which can abstractively generate and instantaneously stitch ad text, matching human quality and achieving real-time ad content customization.  
    \item We record a significant click-through-rate gain of 13.28\% over an extraction-based system as a baseline. Our system is currently deployed at web scale, serving ${\sim}4\%$ of ads shown on Bing search engine.
\end{enumerate}

\section{Ads Generation System}
\label{sec:genMethodology}
Our system for ad content generation and stitching is automated end-to-end as shown in Figure~\ref{fig:DeepGenWorkflow}. Advertisers only need to supply us with their domain names, landing page targeting rules, and a bid for each rule (e.g., bid \$0.5 for URLs containing ``shoes''). Our Search Indexing infrastructure crawls all landing pages under advertiser domain names that match targeting rules and runs the Document Understanding (DU) pipeline to extract textual information as per Section~\ref{sec:Models}.
After that, we run multiple NLG models concurrently. This parallel design enables us to scale modeling horizontally: we can add or remove generation models at will. The models can either generate an ad asset or a full ad copy. For a full ad copy we simply split it into assets. At the end of generation stage, we have many title and description assets generated for each advertiser URL.

\subsection{Baselines}
\label{sec:Models}
\paragraph{Extraction-based systems} The extraction techniques have evolved in Bing Ads over a decade and we consider them a strong industrial baseline in this paper. This baseline can produce title assets of high quality, but it does not perform as well for the longer description assets. For extraction candidates, we leverage parts of the website extracted by Bing DU pipeline, as per example below:
\begin{itemize}[noitemsep]
    \item Page Title - the document title present in metadata; \verb|<|title\verb|>| tag for HTML documents
    \item Visual Headings - the visually emphasized document title present in the document, visible to user
    \item First/Best Body Snippet - first (top-most)/best document body snippet extracted by Bling \citep{xiong2019open}
\end{itemize}
Examples of the above landing page text extracted by DU pipeline can be seen on the left in Figure~\ref{fig:DeepGenWorkflow}. 

\begin{figure}[ht]
\centerline{\includegraphics[width=\columnwidth]{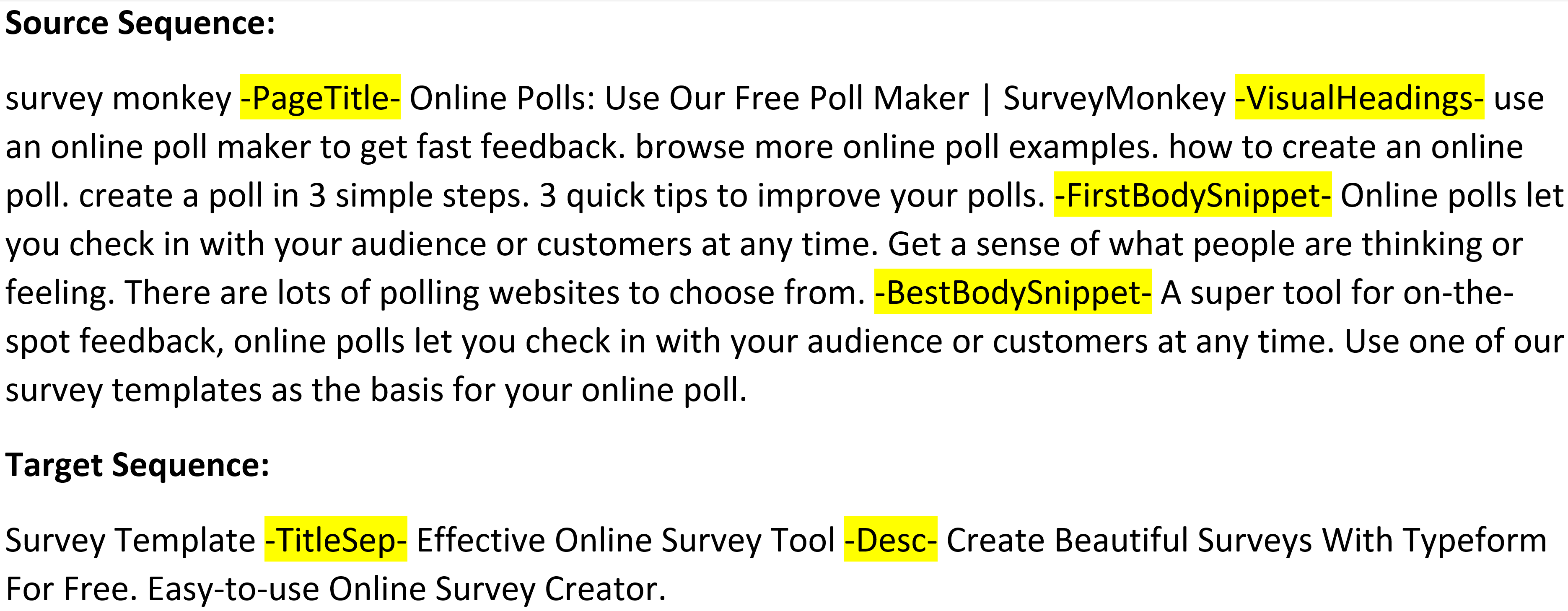}}
\caption{An example of a source and generated target sequence pair for the baseline AdCopy model.}
\label{fig:SrcTgtSeq}
\end{figure}

\paragraph{Abstractive generation baseline} We consider models finetuned directly on advertiser written ad copies as the baseline for abstractive generation approach. We finetune UniLMv2 \citep{UniLMv2} on advertiser-written full ad copies, with learning rate of $5\cdot10^{-5}$. We refer to such models as AdCopy models as they generate one ad copy for each source sequence. See Figure~\ref{fig:SrcTgtSeq} for an example of source/target sequences for this task. Multiple AdCopy models were successfully deployed in production with significant business gains \citep{wang2021reinforcing}. Some best practices we learned are: 1) advertiser-written ads have a very skewed distribution with some advertiser having millions of template generated ads. Therefore we sample the 3000 URLs with the most ad impressions in the past year per advertiser domain, obtaining 3M-5M training examples; 2) validation and test sets randomly split from training set do not work well; they need to be constructed from different advertisers than those in training set to avoid overfitting. We use validation set of size 300K-500K examples and a test set of 30K-50K examples. We use ROUGE1-F1 \cite{lin2004rouge} on validation set to select the best checkpoint during training.

We inference with beam search of size 5 with code optimization, leveraging Einsum operator in cross-attention stage to avoid the encoder cache copy, per the FastSeq \cite{fastseq} implementation. This optimization allows us to increase batch size and brings ~5x speed up in our task. Our generation models can be seen in the center of Figure~\ref{fig:DeepGenWorkflow} in orange color.

\subsection{Factuality Improvement}
\label{sec:Factuality}
To evaluate the quality of generated ads, we mainly rely on human evaluation. For that, we sample a stratified sample of at most 50 examples per domain, and then uniformly subsample 500 -- 1000 examples per human evaluation task. This way, we get an evaluation result from diverse portions of our demand, not letting very large domains dominate. We work with a pool of professional judges, trained to evaluate ads in an unbiased way. We further examine evaluation examples and give feedback to the judges in case there is a misunderstanding of the judgement guidelines. Thus, we evaluate the quality of generated ad texts along the following 4 aspects: 
\begin{itemize}[noitemsep]
    \item Text Quality: evaluates grammar and style, with levels Good, Fair, Bad, Embarrassing, and Not Scorable.
    \item Human Likeness: whether it looks like human-written, with levels Yes and No.
    \item Factuality: whether the generated information is supported by landing page, with levels Yes and No.
    \item Relevance: whether the generated text is relevant to advertiser's business, with levels Yes and No.
\end{itemize}

We define an ad text to be ``Overall Good'' if it gets ``Good'' or ``Fair'' for Text Quality, and ``Yes'' for Human Likeness, Factuality, and Relevance. Refer to Figure~\ref{fig:HitAppUI} in the Appendix~\ref{sec:AppendixA} for an example human judge interface. To be allowed for further A/B testing, the Overall Good Rate needs to be at least 90\% with confidence greater than 97.5\%.

As shown in Table \ref{tab:GoodRates}, our baseline model does not have a significant difference in quality from the advertiser written ads. However, the overall good rate for both is curtailed by lower factuality scores. For example, our AdCopy model can generate popular claims like ``Free Shipping" or ``15\% Discount" which do not exist in the landing page. This is similar to the hallucination issue in abstractive summarization \cite{Filippova2020Controlled,Maynez2020Faithfulness}.

To alleviate the extrinsic hallucinations \citep{maynez-etal-2020-faithfulness} in our ads, we employ phrase-based cross-check filtering. For that, we use a list of potentially erroneous phrases and patterns obtained by studying human evaluation results for our generated ads. Our approach is similar to entity-based filtering per \citet{nan-etal-2021-entity}.

Some cross-check examples are 1) Phrase Check: a list of sensitive or potentially misleading phrases (e.g., ``Free Return'', ``Promo Code: ABC''); 2) Brand Check: brand list compiled from our search engine's knowledge graph  \citep{Noy2019,chai-etal-2021-automatic}; 3) Domain Check: checking patterns like ``xyz.com'' against landing page URL. 

We add the cross check rules at two stages: (1) We filter training data with cross check rules before training (train x-check); and (2) We filter generated text after the inference (infer x-check). Per Table~\ref{tab:GoodRates}, both train x-check and infer x-check improve quality significantly, with the greatest improvement when both are used together.

For an AdCopy model, we do observe that $\sim15\%$ of generated ad copies are filtered during the post-inference cross check. This effect is ameliorated by the fact that we use multiple NLG models, allowing them to backfill each other's coverage. The remaining coverage is backfilled with extraction candidates. Due to this system design, the eventual URL coverage does not suffer from the cross check.

\begin{table*}[ht]
\centering
\begin{tabular}{lccccc}
\hline
\textbf{Technique} & \textbf{Overall} & \textbf{Text Quality} & \textbf{Human Like} & \textbf{Factuality} & \textbf{Relevance}\\
\hline
Advertiser-written & 90.7 $\pm$ 2.1 & 97.9 $\pm$ 1.0 & 98.1 $\pm$ 1.0 & 92.7 $\pm$ 1.9 & 99.0 $\pm$ 0.7 \\
Baseline: AdCopy w/o check & 89.8 $\pm$ 2.2 & 98.8 $\pm$ 0.8 & 98.5 $\pm$ 0.9 & 91.1 $\pm$ 2.1 & 98.9 $\pm$ 0.7 \\ 
\hline
AdCopy w/ train check & \textbf{94.7} $\pm$ \textbf{1.6} & \textbf{99.6} $\pm$ \textbf{0.5}& \textbf{99.0} $\pm$ \textbf{0.7} & \textbf{95.6} $\pm$ \textbf{1.5} & 99.6 $\pm$ 0.5 \\
AdCopy w/ infer check & \textbf{94.4} $\pm$ \textbf{1.9} & 98.8 $\pm$ 0.9 & 98.5 $\pm$ 1.0 & \textbf{95.6} $\pm$ \textbf{1.7} & 98.8 $\pm$ 0.9\\
AdCopy w/ train + infer check & \textbf{96.3} $\pm$ \textbf{1.5} & \textbf{100.0} & \textbf{99.4} $\pm$ \textbf{0.6} & \textbf{97.0} $\pm$ \textbf{1.3} & \textbf{99.7} $\pm$ \textbf{0.4}\\
\hline
\end{tabular}

\caption{A comparison of Ad Copy models (as per Section~\ref{sec:Models}) via human evaluation. 95\% confidence intervals (CI) are reported. Results that outperform advertiser baseline at $p<0.05$ level are \textbf{bolded}. }
\label{tab:GoodRates}
\end{table*}

\begin{table*}[ht]
\centering
\begin{tabular}{lccccc}
\hline
\textbf{Technique} & \textbf{Overall} & \textbf{Text Quality} & \textbf{Human Like} & \textbf{Factuality} & \textbf{Relevance}\\
\hline
Advertiser Title Asset & 98.2 $\pm$ 0.9 & 99.9 $\pm$ 0.2 & 100.0  & 98.4 $\pm$ 0.9 & 100.0\\
Extraction Title Asset & \textbf{99.0} $\pm$ \textbf{0.7} & 99.4 $\pm$ 0.6 & 99.6 $\pm$ 0.5 &  \textbf{99.6} $\pm$ \textbf{0.5} & 100.0\\
Guided Title Asset & 98.1 $\pm$ 0.6 & 99.8 $\pm$ 0.2 & 100.0 & 98.3 $\pm$ 0.5 & 99.6 $\pm$ 0.3 \\
\hline
Advertiser Desc Asset & 98.2 $\pm$ 0.9 & 99.9 $\pm$ 0.2 & 99.9 $\pm$ 0.2 & 98.4 $\pm$ 0.9 & 100.0 \\
Guided Desc Asset & 95.3 $\pm$ 0.9 & 97.6 $\pm$ 0.7 & 98.8 $\pm$ 0.5 & 97.9 $\pm$ 0.6 & 99.2 $\pm$ 0.4 \\
\hline
\end{tabular}

\caption{A comparison of Guided Asset generation model against advertiser written ads and extraction-based titles via human evaluation. 95\% CI are reported. Results better than advertiser baseline at $p<0.05$ level are \textbf{bolded}.}
\label{tab:GoodRatesGuided}
\end{table*}

\subsection{Controllable Generation at Asset-Level}
\label{sec:GuidedGen}

To model diversity explicitly, we build a controllable NLG model to generate multiple ad assets for the same source sequence. We accomplish this is via \textit{control codes}, categorical variables that represent the desired output property and are pre-pended to the model inputs during training and testing, \citet{Keskar2019} and \citet{Ficler2017}. We refer to it as Guided model, as the generation is guided by the control codes. 

We assume each landing page can be advertised along 12 categories for different selling points. Example categories are Product or Service, Advertiser Name or Brand, Location, etc.; they are borrowed from the instructions on the web portal where advertisers create ads. We then use human judges to classify ${\sim}6500$ distinct advertiser-written assets into categories. We finetune BERT-base-uncased \cite{BERT} for asset category classification task and obtain ${\sim}80$\% prediction accuracy, using a random 80/20 split for train/test sets and learning rate of $5*10^{-5}$.

We then inference ad category for each ad asset in the NLG model training set, prepending the resulting category control code as plaintext at the beginning of each NLG source sequence. Thus, we obtain a data set of 6M ad assets (both title and description together) for training the Guided NLG models. Otherwise, our generative modeling decisions align with Section~\ref{sec:Models}. During inference, we evaluate the model on all available categories, by prepending each control code to the landing page information.

Human evaluation results for our Guided NLG model are shown in Table~\ref{tab:GoodRatesGuided}. The overall title asset quality of the Guided model does not have significant difference to that of advertiser-written assets, with Extraction titles outperforming both. The advertiser-written description assets are better, though the overall good rate of Guided model is still well above our quality bar of 90\%. The advantage of Guided model in this case is that it is able to explicitly capture different advertising categories for both title and description. Extraction technique cannot produce good ad descriptions in our experience.

\section{Serving and Customization System}
\label{sec:SystemDesign}

\begin{table}[t]
\centering
\begin{tabular}{l c ccc}
\hline
\textbf{Title Asset} & \textbf{Count} & \textbf{PB}$\downarrow$ & \textbf{SB}$\downarrow$ & \textbf{Dist}$\uparrow$\\
\hline
Advertiser & 18.4 & 13.4 & 71.0 & 45.3 \\
Generated & 24.4 & 6.7 & 41.0 & 66.6 \\
Generated + DPP & 14.2 & \textbf{4.5} & \textbf{25.3} & \textbf{80.5} \\
\hline
Guided & 13.3 & 7.8 & 33.6 & 74.9 \\
Ensemble & 12.1 & \textbf{5.8} & \textbf{31.2} & \textbf{77.0} \\
\hline
Guided + DPP & 7.8 & 5.0 & 18.3 & 86.9 \\
Ensemble + DPP & 7.7 & \textbf{3.6} & \textbf{17.0} & \textbf{88.3} \\
\hline
\end{tabular}
\caption{Averaged results of the diversity evaluation on English title assets. For PairwiseBLEU (PB) and SelfBLEU (SB) scores, lower is better, for Distinct N-gram (Dist) scores, higher is better. Average count of title assets per URL (Count) is also reported. Differences of over 1 point are \textbf{bolded}. Ensemble here is for an ensemble of AdCopy models. Generated assets include the combination of Guided, Ensemble, and Extraction titles.}
\label{tab:TitleDiversity}
\end{table}

\subsection{Diverse Selection}
\label{sec:diverseSelection}
At this stage, we aim to select a semantically diverse subset of $T$ title and $D$ description assets for each URL to send to online serving components. By selecting a subset of ad texts, we aim to both reduce the load on the ad serving system, as well as improve diversity of the generated texts.  We use CDSSM \cite{cdssm} model, trained on web search logs, to map each text asset to a dense vector, such that the ad texts with high degree of semantic similarity will map to representations with higher cosine similarity (i.e., closer in the embedding space) to one another. Then, we sample a diverse subset of points in the CDSSM embedding space with k-DPP maximum a posteriori inference algorithm as per \citet{Chen2018DPP}, stopping after we select $T$ titles or $D$ descriptions. Refer to Figure~\ref{fig:DeepGenWorkflow} (bottom middle) for an example of removing semantic duplicates in such fashion.

We use PairwiseBLEU (PB) \citep{Shen2019Mixture}, SelfBLEU (SB) \citep{Texygen}, and Distinct N-gram (DistN) \citep{xu-etal-2018-diversity} scores to evaluate the diversity of the title assets before and after k-DPP diverse sampling. We calculate the average diversity metrics for ${\sim}2000$ EN URLs randomly sampled from a stratified sample of 50 URLs/domain. Since all instances of each metric show similar trends, we follow suit with \citet{Tevet2020Evaluating} and average each metric over different N-gram options. Refer to Table~\ref{tab:TitleDiversity} for diversity score details.

We find that generated title assets are more diverse than the ones provided by the advertiser in general. In addition, k-DPP helps further increase the asset diversity. We also compare title assets from the Guided model with those from an ensemble of AdCopy models. We find that the Guided model by itself can generate title assets in similar quantity and with similar diversity as the ones produced by several AdCopy models combined, trained as per the NLG baseline method in Section~\ref{sec:Models} on different versions of training data. 

\subsection{Real-Time Stitching}
\label{sec:stitching}
The diversified ad assets are then ingested into the online serving infrastructure. At query time, we stitch together a customized ad copy, optimizing for the auction win rate\footnote{Auction is the final stage to decide which ads will be displayed. Auction win rate is the probability of an ad winning an auction. Ads with better quality and CTR have a better chance to win the auction.} (with some level of exploration). From our domain knowledge, the earlier asset positions (e.g., Title 1) influence the ad auction result more than the later ones (e.g., Description 2), as shown in Figure \ref{fig:AdComponentsXbox}. Thus, we perform a greedy sequential selection and consider $T+(T-1)+(T-2)+D+(D-1)$ permutation options. For example, we first select asset for Title 1 position from $T$ title assets, and then select asset for Title 2 position from the remaining $T-1$ title assets.

We use a logistic regression (LR) model to score each asset position: Title 1, Title 2, Title 3, Description 1, Description 2. We use features from ad auction log like string hash, length, unigrams and bigrams from asset texts. We also cross these with the query text to a total sparse feature dimensionality per position of ${\sim}4$B. The LR model learns the probability of winning the auction for a given ad copy. It is continuously trained daily, using ${\sim}10$B data examples from the previous day's log for training with batch size as $1000$ and learning rate as $0.02$, and $\sim{300}$M examples from current day's log for validation. 

We include an exploration mechanism to allow newly added assets to be shown to  users and to de-bias the model. Due to sequential nature of our stitching process, we model exploration as a sequential contextual bandit (CB) problem. At each asset position, the CB uses the LR score and the gradient sum of LR features as a heuristic for the trial count \citep{Mcmahan2013} to select an asset using Thompson Sampling strategy \citep{Agrawal2017NearOptimal}. As a result, we sample from a total of $T + (T-1) + (T-2) + D + (D-1)$ Beta distributions to stitch together an ad copy.


\begin{table*}[t]
\centering
\begin{tabular}{|c|c|c|l|}
\hline
\textbf{Metric} &\textbf{Exp. 1} & \textbf{Exp. 2} & \textbf{Explanation}\\
\hline
Days & 5 & 10 & Number of days for the experiment. \\
Traffic\% & 5.0 & 10.0 & Percentage of the Bing user traffic allocated for the experiment. \\
$\Delta$RPM\% & +\textbf{24.87} & +\textbf{10.65} & Revenue (USD) from 1000 Search Results Page Views (SRPVs). \\
$\Delta$IY\% & +\textbf{11.87} & +\textbf{14.43} & Average number of ads shown per page.\\
$\Delta$CTR\% & +\textbf{13.28} & -0.19 & Proportion of ads clicked from ads shown.\\
$\Delta$QBR\% & +\textbf{5.27} & +\textbf{1.82} & Proportion of ad clicks that resulted in a back-click within 20 sec.\\
\hline
\end{tabular}
\caption{A summary of the business metrics from A/B tests performed on DSA ad traffic. Results statistically significant at $p<0.05$ level are \textbf{bolded}.}
\label{tab:ABTest}
\end{table*}

\section{A/B Testing}
\label{sec:ABTests}
DeepGen is deployed globally to serve Dynamic Search Ads (DSA), which accounts for $\sim4\%$ of all Bing Ads displayed globally. In A/B testing, we split the production user traffic randomly between the treatment experiment that enables the proposed experimental techniques and the control experiment that uses existing production techniques. We use 10\% of production traffic for the control experiment. We use the difference in business metrics between two experiments to decide if treatment is effective.

Two key business metrics are Revenue Per Mille (RPM) – revenue per every thousand search result page views (SRPV) and Quick Back Rate (QBR) – the rate of users clicking the back button after clicking on an ad, which is a proxy for user dissatisfaction (lower QBR is  better). RPM is driven by Impression Yield (IY, number of ads shown divided by number of search result page views) and Click-Through Rate (CTR, number of clicks divided by total number of ads displayed). Usually there is a trade-off between RPM (revenue) and QBR (user satisfaction). DeepGen increases CTR (proportion of ads clicked from ads displayed) and IY (number of ads displayed per page), thus also increasing ad revenue. We do so by generating high-quality ads that are customized to the user. We avoid sacrificing user or advertiser satisfaction by ensuring the ads to be faithful to the landing page.

In Exp. 1, we compare DeepGen (treatment) against the extraction system (control). As shown in Table \ref{tab:ABTest}, we observe strong RPM (revenue) gain, driven by both IY  and CTR, which means that personalized ad copies generated by DeepGen are more likely to win the auction as well as to be clicked by the user. In this experiment, we record a \textit{13.28\% CTR gain}. We acknowledge the increase in QBR (user dissatisfaction), which could be attributed to the still higher factuality of the extraction system, as shown in Table \ref{tab:GoodRatesGuided}. 

We use Exp. 2 as an ablation for real-time customization. DeepGen is used in both treatment and control, but we replace real-time stitching with pre-computed stitching in control. For this experiment, we build a separate model to stitch assets into multiple ad copies offline, and only \textit{rank} the pre-stitched ad texts during query time (online). There is significant RPM (revnue) gain, though it is mainly driven by IY but not CTR. This may suggest that online stitching has a higher chance of winning the auction as it covers much larger permutation space than the offline stitching. But for those ad copies that did win an auction, they have similar attractiveness to the user whether stitched online or offline. This experiment shows online stitching to be an integral part of our system.

Thus, DeepGen increases revenue by generating high-quality ads customized to the user while being mindful of user satisfaction by ensuring the ads to be faithful to the landing page.

\section{Related Work}
\label{sec:Literature}
The early automated content generation approaches focused on template-based ad text generation \citep{Bartz2008Ads, Fujita2010Gen, Thomaidou2013Snippet}. These approaches have potential to suffer from ad fatigue \citep{Abrams2007AdFatigue}.

More recently, deep Reinforcement Learning (RL) was shown effective for ad text generation \citep{Hughes2019RL, Kamigaito2021AnES, wang2021reinforcing}, using a \textit{general} attractiveness model as a reward policy and yielding up to 7.01\% observed CTR gain per \citet{Kamigaito2021AnES}. CTR is an important metrics, as reflects the relevance of an ad from user's perspective \citep{Yang_2022}.

Product headline generation is a closely related direction of work, where a single headline is generated to advertise a line of related products, based on each product's advertiser-written title. \citet{kanungo-etal-2021-ad} use BERT-large \citep{BERT} encoder finetuned for generation with UniLM-like masked attention, as per \citet{UNILM}, optimized using a self-critical RL objective, as per \citet{Hughes2019RL}. \citet{Kanungo2022} further produce SC-COBART by finetuning a BART model, using control codes, as per \citet{Keskar2019}, for bucketized CTR and length of a headline, optimized with a mixture of MLE and self-cricial RL objectives. SC-COBART improves estimated CTR by 5.82\% over their previous work \citep{kanungo-etal-2021-ad}. 

In another line of work, product descriptions are generated either with templates \citep{wang-etal-2017-statistical}, pointer-generator encoders \citep{Tao2019Pointer}, commonsense knowledge-base guidance \citep{KOBE, CHASE}, or CVAEs \citep{Shaeo2021Apex}, yielding up to 13.17\% CTR gain in A/B test per \citet{Shaeo2021Apex}. 

\section{Conclusion}
\label{sec:conclusion}
In this work, we present an automated end-to-end search advertisement text generation solution. We employ deep NLG modeling for ad content generation and diverse selection. We leverage real-time LR rankers for content stitching. The generation techniques provide us a rich source of high-quality ad content, which performs strongly against human and extraction baselines. We further apply diverse selection via semantic embedding, which allows us to surpass human content diversity, while ensuring the system's scalability. Finally, we use real-time ranking to stitch not just attractive, but a truly \textit{customized} ad for each user based on query and search intent. The system combines several NLP approaches to provide a cutting edge solution to automated ad generation and showcases an significant CTR gain over an extraction baseline.

\bibliography{main}
\bibliographystyle{acl_natbib}

\appendix

\section{Ad Text Quality Judgement UI}
\label{sec:AppendixA}
\begin{figure}[ht]
\centerline{\includegraphics[width=\columnwidth]{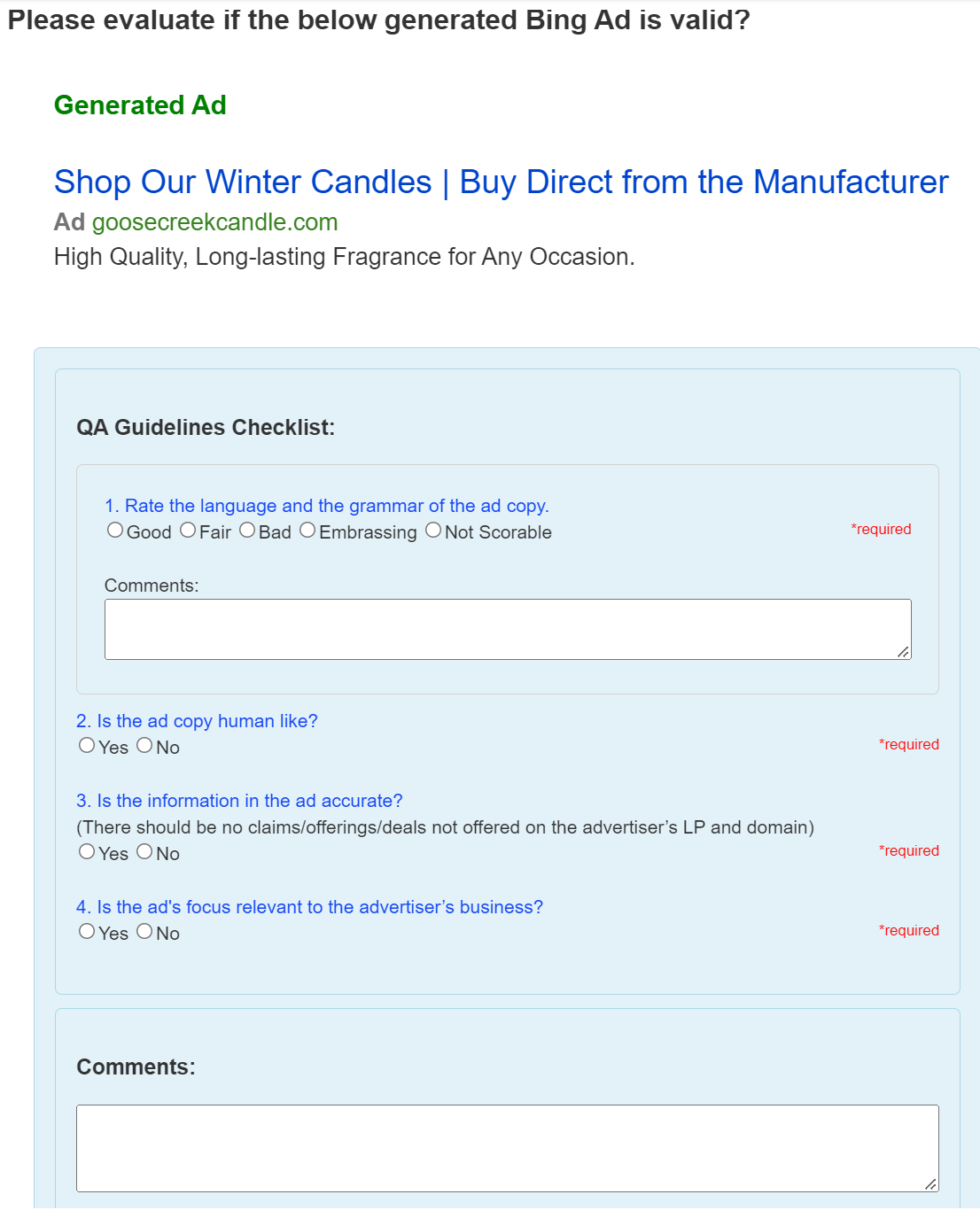}}
\caption{User interface for human ad quality evaluation.}
\label{fig:HitAppUI}
\end{figure}

\end{document}